\documentclass[runningheads]{llncs}
\usepackage{graphicx}

\usepackage{xcolor}
\usepackage{hyperref}
\usepackage{bbding}
\usepackage{multirow}
\usepackage{footnote}
\usepackage[normalem]{ulem}
\useunder{\uline}{\ul}{}
\usepackage[caption=false, font=scriptsize]{subfig}

\usepackage{chngcntr}
\counterwithout{figure}{chapter}
\counterwithout{table}{chapter}

\begin{document}
%
\title{\large{Evaluating Explainable Methods for Predictive Process Analytics: A Functionally-Grounded Approach}}
\titlerunning{Evaluating Explainable Methods for Predictive Process Analytics}
%
\author{Mythreyi Velmurugan, 
Chun Ouyang, 
Catarina Moreira, 
Renuka Sindhgatta 
}
\authorrunning{M. Velmurugan et al.}
%
\institute{Queensland University of Technology, Brisbane, Australia
\email{\{m.velmurugan,c.ouyang,catarina.pintomoreira,renuka.sr\}@qut.edu.au}}
\maketitle              
\begin{abstract}
Predictive process analytics focuses on predicting the future states of running instances of a business process. 
While advanced machine learning techniques have been used to increase accuracy of predictions, the resulting predictive models lack transparency. Current explainable machine learning methods, such as LIME and SHAP, can be used to interpret black box models. However, it is unclear how fit for purpose these methods are in explaining process predictive models. In this paper, we draw on evaluation measures used in the field of explainable AI and propose functionally-grounded evaluation metrics for assessing explainable methods in predictive process analytics. We apply the proposed metrics to evaluate the performance of LIME and SHAP in interpreting process predictive models built on XGBoost, which has been shown to be relatively accurate in process predictions. We conduct the evaluation using three open source, real-world event logs and analyse the evaluation results to derive insights. The research contributes to understanding the trustworthiness of explainable methods for predictive process analytics as a fundamental and key step towards human user-oriented evaluation.

\end{abstract}
\section{Introduction}\label{intro}

Modern data analytics underpinned by machine learning has enabled data-led decision making in the information era. 
Predictive process analytics, as an important branch of data analytics, focuses on applying predictive analytics to forecast future states of business process execution, for example, to predict the most likely outcome of a running process instance~\cite{Teinemaa2019}. 
While advanced machine learning techniques have been used to increase accuracy of process predictions, the resulting predictive models become less transparent due to their sophisticated internal representations. 
These are often called ``black box'' models, and methods and techniques have been proposed in machine learning to explain such models, forming a new research theme known as explainable AI (XAI)~\cite{Guidotti2018}.

Several recent studies in predictive process analytics have attempted to apply existing explainable methods in XAI to interpret black box models used to compute process predictions. 
For example, LIME and SHAP have previously been used to interpret process predictive models built on complex machine learning algorithms~\cite{Galanti2020,Rizzi2020,Sindhgatta2020}. 
While it is important to understand how well the explanations generated by an explainable method may be received by end users, it is essential to study how fit for purpose such an explainable method is when being used to explain predictive models for process predictions, and 
how the characteristics of datasets and underlying machine learning techniques affect the quality of explanations generated. 
Assessing the latter, essential aspect of an explainable method is referred to as a functionally-grounded evaluation in XAI~\cite{DoshiVelez2017}.

In this paper, we identify and draw on evaluation measures used in XAI, and propose functionally-grounded evaluation metrics for assessing explainable methods in the context of predictive process analytics. 
We apply the proposed metrics to evaluate the performance of LIME and SHAP in interpreting process predictive models built on XGBoost, which has been shown to be most accurate in process outcome predictions~\cite{Teinemaa2019}. 
We conduct the evaluation using three open source event log datasets that record the execution of three real-life business processes in different industries, and analyse the evaluation results to discover insights and knowledge. 
The research contributes to understanding the trustworthiness of explainable methods for predictive process analytics as a fundamental and key step towards human user-oriented evaluation.


The paper is organised as follows. 
It provides a brief overview of XAI, including XAI evaluation, and state-of-the-art explainable predictive process analytics, and highlights the gaps that this research attempts to address in Sect.~\ref{background}. 
This is followed by definition of the proposed evaluation metrics in Sect.~\ref{metrics}. 
Details of the conducted evaluation, elaborated in terms of the design of experiments, real-life datasets, and analysis and findings drawn from the experiments, are presented in Sect.~\ref{evaluation}. 
Finally, Sect.~\ref{concl} concludes the paper with an outlook for future work. 


\section{Background and Related Work}\label{background}



\subsection{Explainable AI}\label{xai}
While more complex algorithms often produce more accurate results, they are harder for a human to understand the workings of, becoming a ``black box'', requiring interpretation~\cite{Guidotti2018}. 
XAI methods are typically classified into two categories: \textit{interpretable models} and \textit{post-hoc interpretation}. 
Interpretable (or ``white box'') models are those that are immediately interpretable by a human (e.g.\ decision trees), though such models are often simpler models and may have reduced predictive power~\cite{Guidotti2018}. 
Post-hoc interpretation refers to the interpretation of a model (global explanation), a data neighbourhood or a prediction (local explanation) created after the predictive model has been generated~\cite{Guidotti2018}. 
Two popular black-box-model-agnostic, local explanation methods in literature are LIME and SHAP. 
LIME determines the importance of all features in an input by perturbing the dataset to create a surrogate linear model that captures the black box model's behaviour at a specific neighbourhood~\cite{Ribeiro2016}. 
SHAP uses a game theoretic approach to assign a value, known as SHAP value, to each feature at the instance level, describing its contribution to the final output (e.g.\ a prediction)~\cite{Lundberg2017}.

\subsection{Evaluating Explanations}
There are a number of ways to evaluate explanations, many of which depend on the purpose of the explanation and the explainee~\cite{Mohseni2020}. A three-level system of evaluation is proposed in~\cite{DoshiVelez2017}, which comprises of: 
\begin{itemize}
    \item \textit{Application-Grounded}: Evaluating explanations in context with end users;
    \item \textit{Human-Grounded}: Evaluating explanations with laypeople doing simple or simplified tasks; and
    \item \textit{Functionally-Grounded}: Using functional tasks without user-based evaluation. An example of such a measure would be computational efficiency~\cite{Mohseni2020}.
\end{itemize}

Two common functional measures of explainability in literature are \textit{stability} and \textit{fidelity}.

\subsubsection{Stability of Explanations}\label{stability} 
Explanation stability is defined as the level of similarity between explanations for similar or identical instances~\cite{Visani2020}. Stability of methods or generated explanations that focus on feature importance attribution can be evaluated in several ways, including measuring the stability of a returned subset of features (\textit{stability by subset}) and the stability of the measure of importance provided to each feature (\textit{stability by weight})~\cite{MohanaChelvan2016}. 
Several measures of stability for XAI have been proposed, but many are specific to a single explainable method, such as stability indices for LIME~\cite{Visani2020}. However, measures proposed to assess the stability of feature selection algorithms can be adapted to evaluate explanation stability where explanations involve feature attribution. These measures typically use sets of features or feature weights, and measure the stability of the sets by averaging the pairwise similarity for each possible pair in the set~\cite{MohanaChelvan2016}. 

\subsubsection{Fidelity of Explanations}\label{fidelity}
Explanation fidelity is generally defined as how faithful the explanation is in relation to the black box. Two ways of measuring fidelity are defined in~\cite{Messalas2019}: external and internal fidelity. \textit{External fidelity} measures the similarity of decisions made by a surrogate model or interpretation of a black box and the black box itself~\cite{Messalas2019}. While this approach can be used to measure how often a black box and its interpreter agree, it does not measure how well the interpreter mimics the black box's decision-making process -- defined as \textit{internal fidelity}~\cite{Messalas2019}. Approaches that have been used to measure internal fidelity include: 
\begin{itemize}
    \item Creating explanations for the black box and surrogate models using an existing explainable method to determine how often the explanations concur~\cite{Messalas2019};
    \item Removing or changing features identified by the interpreter and comparing the changes in prediction probability of the black box~\cite{Du2019,Kindermans2019}; and
    \item Creating explanations for a white box model to see how well the explanations match the decision-making of the white box~\cite{Ribeiro2016} (though this does not necessarily imply that the interpreter will perform as well as with a more complex model). 
\end{itemize}

\subsection{Explainable Predictive Process Analytics}
State-of-the-art explainable predictive process analytics have generally attempted to use existing explainable methods in XAI. For example, the use of LIME and SHAP to evaluate and improve black box models~\cite{Rizzi2020,Sindhgatta2020} has been explored, and SHAP has been used to create explainable dashboards for informed decision-making~\cite{Galanti2020}.  A method of interpreting process predictive models underpinned by neural networks using layer-wise relevance propagation 
is presented in~\cite{Weinzierl2020a}. In contrast to these, a semi-interpretable, process-aware approach was proposed for remaining time predictions, where predictive models were trained for every activity and decision point in the process, and a prediction was made at each step to better understand how each activity and decision point contributed to the remaining time~\cite{Verenich2019a}.

\subsection{Problem Statement}
There has been few attempts in explainable predictive process analytics to assess the quality of explanations. 
With existing XAI evaluation methods, many are specific to a particular explainable method~\cite{Visani2020} or prediction problem~\cite{Du2019,Kindermans2019}. 
In particular, methods of testing fidelity often apply ablation, in which the most influential features are removed from the input~\cite{Du2019}, or perturbing the data to effect a semantically meaningless transformation to determine the limits of interpreter in withstanding meaningless change~\cite{Kindermans2019}. However, these are typically applied to text or image data, and would not be appropriate for tabular data, such as event logs. 
This has motivated us to draw on existing work to build extensible and functionally-grounded evaluation metrics and design associated methods to assess explainable predictive process analytics.

\section{Functionally-Grounded Evaluation Metrics}
\label{metrics}

We take a functionally-grounded evaluation approach, using functionality-based measures without human evaluation~\cite{DoshiVelez2017}. More specifically, we consider explanation quality through stability and fidelity of provided explanations. It is important to note that the definition of each measure in this section focuses on the (local) explanations at the process instance level for any given event log dataset. Evaluation of each measure is based on the fact that explanations are generated multiple times for each individual process instance. The overall evaluation of such a measure over the entire event log can be calculated as the average total of the evaluations of that measure of all individual process instances in the log. 

\subsection{Explanation Stability}

Explanation stability will be evaluated in terms of the stability of subsets of most important features and the stability of feature weights. 

A \textit{stability by subset} measure, as proposed in~\cite{Nogueira2018}, is based on the presence or absence of each feature across a number of subsets. As such, we calculate the stability of feature subsets ($\phi(\mathcal{Z})$) for a single process instance in an event log as follows: 

\begin{equation}\label{eq:subset-stability}
\phi(\mathcal{Z})= 1 - \frac{\frac{1}{d}\sum^d_{i=1}s^2_{f_i}}{\frac{\overline{k}}{d}(1-\frac{\overline{k}}{d})}
\end{equation}

where:
\begin{itemize}
    \item $d$ = number of features encoded from event attributes in the log 
    \item $M$ = number of explanations generated for the process instance
    \item $k$ = number of most relevant features, where relevance or level of importance is determined by an explanation generated for the process instance
    \item $\overline{k}$ = average number of features selected across all $M$ explanations for the process instance
    \item $s^2_{f_i}$ = sample variance of the presence of feature ${f_i}$ across all $M$ explanations for the process instance
    \item $\mathcal{Z}$ = binary matrix of size $M$ x $d$. Each row of the binary matrix represents a feature subset, where a 1 at the $i^{th}$ position means feature $f_i$ has been selected and a 0 means it has not been selected.
\end{itemize}
This measure is bounded between 0 and 1, where 0 indicates no similarity in the feature subsets, and 1 indicates that all subsets are identical. 

Next, Pearson's correlation coefficient is generally used to measure stability of feature weights in feature selection algorithms~\cite{MohanaChelvan2016}, but this is a measure of the general trend of importance and does not calculate the degree by which a feature's weight may vary. As such, we specify the measure \textit{stability by weight} based on the statistical measure of relative variance, and calculate the stability of feature weights ($\phi(\mathcal{W})$) for a single process instance in an event log as follows: 

\begin{equation}\label{eq:weight-stability}
\phi(\mathcal{W}) = 1-\frac{1}{d}\sum^d_{i=1}\frac{\sigma^2_{w_i}}{|\mu_{w_i}|}
\end{equation}

where:
\begin{itemize}
    \item $d$ = number of features encoded from event attributes in the log 
    \item $M$ = number of explanations generated for the process instance
    \item $\mu_{w_i}$ = mean of the weights of feature $f_i$ across all $M$ explanations for the process instance
    \item $\sigma^2_{w_i}$ = variance of the weights of feature $f_i$ across all $M$ explanations for the process instance 
    \item $\mathcal{W}$ = matrix of size $M$ x $d$. Each row of the matrix records the weight of each feature as quantified by an explanation
\end{itemize}
This measure also has an upper bound of 1 (indicating perfect stability), but no lower bound. 

\subsection{Internal Fidelity}
We evaluate the \textit{internal fidelity} of explanations, 
as we are interested in the fidelity of the interpreter's decision-making processes, not the decision (i.e.\ prediction) itself. Error functions have previously been applied to quantify internal fidelity, averaged out over the size of a dataset~\cite{Du2019}. As such, the mean absolute percentage error (MAPE) can be used to measure the fidelity of explanations, 
and we calculate the fidelity ($\mathcal{F}$) of the interpreter for a single process instance in an event log as follows: 



	
\begin{equation}\label{eq:overall-fidelity}
\mathcal{F} = \frac{\sum^{|X'|}_1\frac{|Y(x)-Y(x')|}{Y(x)}}{|X'|}
\end{equation}
where:
\begin{itemize}	
\item $x$ = original feature vector for the process instance 
\item $X'$ = Set of perturbations for $x$ and $x'\in X'$
\item $Y(x)$ = Prediction probability given input $x$
\item $Y(x')$ = Prediction probability given input $x'$
\end{itemize}

This measure is naturally bounded by the fact that prediction probabilities fall between 0 and 1.
\section{Evaluation and Analysis}\label{evaluation}

\subsection{Design of Experiments}\label{setup}


Process instance outcomes were chosen as the prediction target since outcome classification is one of the most common predictive process analytics problems. 
XGBoost was the machine learning algorithm used as overall it is the most accurate for outcome prediction~\cite{Teinemaa2019}. 
The XGBoost classifiers were trained on different data encoding and bucketing methods. The following combinations of bucketing and encoding methods were used:
\begin{itemize}
    \item Aggregate encoding for dynamic attributes with prefix-length bucketing;
    \item Index-based encoding for dynamic attributes with prefix-length bucketing; and
    \item Aggregate encoding for dynamic attributes with no bucketing.
\end{itemize}

In the ``no bucketing'' method, all data is compiled as one (i.e. a single bucket) and a single classifier is trained on this bucket. When prefix-length bucketing is used, data is grouped (bucketed) based on shared prefix length (the number of activities that have already been completed in a process instance), and a classifier is trained for each bucket. For example, in a dataset with process traces ranging from prefix length of 1 to 40, forty classifiers will be trained. 

Static encoding is the most effective way to encode static attributes (i.e. attributes that do not change over the course of the process trace, such as diagnosis or patient age) and is used in all combinations. In this method, numeric attributes are encoded as-is and one-hot encoding is used for categorical attributes. Aggregate and index-based encoding are applied for dynamic attributes (i.e. attributes that change over the course of the process instance, e.g., activity completed or time taken at each step). If using aggregate encoding, the frequency or number of occurrences of each categorical attribute is used as a feature, while the minimum, maximum, mean and standard deviation of feature values are all used as features for each numeric attribute. If index-based encoding is used, numeric attributes are encoded as-is and categorical attributes are one-hot encoded at each index (prefix in the process trace).

SHAP and LIME, two popular post-hoc interpretation methods, are chosen as the explainable methods to be evaluated, as they have previously been used in predictive process analytics literature~\cite{Galanti2020,Rizzi2020,Sindhgatta2020}. 

As a summary, each combination of the above bucketing, encoding and explainable methods were evaluated.

\subsubsection{Evaluating Stability}
Ten explanations were generated for each instance in a sample of the test set (i.e. $M = 10$ for each instance), and the mean of the explanation stability for all features in an instance denotes the stability for that instance. 
When computing stability by subset, the features selected in each explanation are ranked based on the importance value attached to that feature, and the ten most influential features are chosen as the subset to be measured using Equation~\ref{eq:subset-stability} (where $k = 10$ and $\overline{k} = 10$). Equation \ref{eq:weight-stability} is used to measure stability by weight, and the quantified feature importance provided in each explanation for all features (not just a subset) are used to calculate stability. Given an event log dataset, the stability scores were averaged for each method combination to indicate the overall stability for that combination. 

\subsubsection{Evaluating Fidelity} 
The evaluation focuses on the internal fidelity of the chosen explainalbe methods. 
Of the three internal fidelity evaluation approaches outlined in Sect.~\ref{fidelity}, we will use the second one. The first approach is intended to assess the fidelity of custom-made surrogate models, not existing methods, and so is not suitable for the aims of this paper. The third one attempts to assess the fidelity of the explanation method by using a white box model as the model to be explained, but this does not necessarily imply that the explainable method will be as faithful for a more complex black box model. As such, changing or removing features based on the explanation to determine fidelity is the most appropriate approach here. 

Next, an ablation approach to measuring internal fidelity will not hold for tabular data like event logs, particularly when using XGBoost which automatically imputes missing data. As such, instead of removing features, a perturbation strategy is considered more effective to measure internal fidelity when using event logs. Again, ten explanations were generated for each instance -- to mitigate the effects of any instability -- and the top 10\% of features 
that occurred most commonly in the explanations were identified. For each feature, LIME presents the feature value or feature value distribution which affected the black box's prediction. For example, an explanation including ``1 $<$ Activity\_A $<$ 3'' indicates that the occurrence of activity~$A$ more than once, but fewer than three times was influential on the result. As SHAP presents only the feature's influence on the end result, feature value distributions based on the SHAP value of that feature had to be generated (i.e. what feature values produce that or similar SHAP values in the entire test set). For example, if ``Activity\_A'' has a SHAP value between 0.5 and 0.6 only when the feature value is between 1 and 3, this would be the distribution attributed to a SHAP value of 0.54. Using these distributions, for each instance:
\begin{enumerate}
    \item A prediction using input vector $x$ was generated, along with the prediction probability for the predicted class ($Y(x)$)
    \item For each feature to be perturbed, a new, uniform distribution outside of the existing distribution was created to draw new feature values from
	\item For each feature to be perturbed, a new value was randomly sampled from the new distribution to replace the original value for that feature to create the perturbed feature vector $x'$
	\item The prediction probability for the originally predicted class was determined for input $x'$ ($Y(x')$) and the difference between $Y(x)$ and $Y(x')$ was computed
\end{enumerate}
Each instance was perturbed ten times, and the differences in prediction probability were used to calculate the MAPE for each instance (see Equation~\ref{eq:overall-fidelity}). 

The source code implementing the proposed evaluation metrics in Sect.~\ref{metrics} and the above approach for design of experiments as well as the associated experiment results are available at:  \url{https://git.io/JIYtH}.

\subsection{Datasets}\label{datasets}
We use three open-source, real-world event logs. All three were used in a predictive process monitoring benchmark on outcome prediction \cite{Teinemaa2019}, and are varied in the amount of cases, types of attributes present and context as shown in Table~\ref{tab:dataset description}. 
We follow the preprocessing, bucketing and encoding methods used in \cite{Teinemaa2019} to prepare the event logs for training, testing and interpreting. 

\begin{table}[h!]
\centering
\caption{A summary of statistics of three event log datasets}
\vspace*{-.25\baselineskip}
\label{tab:dataset description}
\scriptsize{
\begin{tabular}{|c|l|l|l|l|}
\hline
\multicolumn{2}{|c|}{\textbf{Event Log}} &
  Production\footnotemark &
  \begin{tabular}[c]{@{}l@{}}Sepsis   \\ Cases\footnotemark\end{tabular} &
  BPIC2012\footnotemark \\ \hline
\multicolumn{2}{|c|}{\textbf{Description}} &
  \begin{tabular}[c]{@{}l@{}}A manufacturing \\ process\end{tabular} &
  \begin{tabular}[c]{@{}l@{}}Hospital event log \\ showing sepsis cases\end{tabular} &
  \begin{tabular}[c]{@{}l@{}}Loan application \\ process\end{tabular} \\ \hline
\multicolumn{2}{|c|}{\textbf{No. of Cases (before encoding)}} &
  220 &
  782 &
  4,685 \\ \hline
\multicolumn{2}{|c|}{\textbf{Proportion of Positive Cases}} &
  55.0\% &
  16.0\% &
  53.4\% \\ \hline
\multicolumn{2}{|c|}{\textbf{Maximum Prefix Length}} &
  23 &
  29 &
  40 \\ \hline
\multicolumn{2}{|c|}{\textbf{Prefix Lengths Used}} &
  1 -- 20 &
  1 -- 25 &
  1 -- 25 \\ \hline
\multirow{4}{*}{\textbf{\begin{tabular}[c]{@{}c@{}}Feature \\ Vector \\ Shape\end{tabular}}} &
  \textbf{\begin{tabular}[c]{@{}l@{}}Single Bucket \& \\ Aggregate Encoding\end{tabular}} &
  162 &
  274 &
  134 \\ \cline{2-5} 
 &
  \textbf{\begin{tabular}[c]{@{}l@{}}Prefix-length \\ buckets \& \\ aggregate encoding\end{tabular}} &
  \begin{tabular}[c]{@{}l@{}}Min: 137\\ Max: 156\end{tabular} &
  \begin{tabular}[c]{@{}l@{}}Min: 153\\ Max: 218\end{tabular} &
  \begin{tabular}[c]{@{}l@{}}Min: 43\\ Max: 134\end{tabular} \\ \cline{2-5} 
 &
  \textbf{\begin{tabular}[c]{@{}l@{}}Prefix-length buckets \& \\ Index-Based Encoding\end{tabular}} &
  \begin{tabular}[c]{@{}l@{}}Min: 100\\ Max: 844\end{tabular} &
  \begin{tabular}[c]{@{}l@{}}Min: 147\\ Max: 535\end{tabular} &
  \begin{tabular}[c]{@{}l@{}}Min: 11\\ Max: 1654\end{tabular} \\ \cline{2-5} 
\hline
\end{tabular}
}
\vspace*{-.75\baselineskip}
\end{table}

\footnotetext[1]{https://doi.org/10.4121/uuid:68726926-5ac5-4fab-b873-ee76ea412399} 
\footnotetext[2]{https://doi.org/10.4121/uuid:915d2bfb-7e84-49ad-a286-dc35f063a460}
\footnotetext[3]{https://doi.org/10.4121/uuid:3926db30-f712-4394-aebc-75976070e91f} 

The Production dataset, derived from a manufacturing process, has the fewest cases and a substantial number of static and dynamic attributes (though more dynamic than static). Cases in which at least one work order is rejected is classified as a positive (deviant) case and around 55\% of the cases in this dataset are positive. The Sepsis Cases dataset records the pathway taken by a patient through a hospital and a positive (deviant) case is one where the patient returns to the ER. This dataset is highly unbalanced (16\% positive cases) and contains considerably more static than dynamic attributes, which will result in comparatively longer feature vector lengths when using aggregate encoding, but comparatively shorter feature vector lengths at higher prefix lengths when using index-based encoding. The training dataset was balanced through downsampling, but the testing data remains unbalanced. The BPIC2012 event log follows a loan process and a positive (deviant) case is one in which the loan isn't accepted (around 53.4\% of the cases). This event log contains one (numerical) static attribute and a number of dynamic attributes, many of which are categorical. This will result in a comparatively smaller feature vector when using aggregate encoding, but the feature vector size will increase considerably as prefix length increases when using index-based encoding.
\subsection{Results and Observations}\label{results}


\begin{table}[b!]
\vspace*{-.5\baselineskip}
\centering
\caption{Overall stability by subset results for each dataset}
\vspace*{-.25\baselineskip}
\label{tab:subset-stability-results}
\scriptsize{
\begin{tabular}{ll|c|c|c|}
\cline{3-5}
 &
   &
  \multicolumn{1}{l|}{\textbf{Production}} &
  \multicolumn{1}{l|}{\textbf{Sepsis Cases}} &
  \multicolumn{1}{l|}{\textbf{BPIC 2012}} \\ \hline
\multicolumn{1}{|l|}{\multirow{2}{*}{\textbf{\begin{tabular}[c]{@{}l@{}}Single bucket\\ Aggregate Encoding\end{tabular}}}} &
  LIME &
  0.72 &
  0.20 &
  0.81 \\ \cline{2-5} 
\multicolumn{1}{|l|}{} & SHAP & {\ul \textbf{1.00}} & {\ul \textbf{1.00}} & {\ul \textbf{1.00}} \\ \hline
\multicolumn{1}{|l|}{\multirow{2}{*}{\textbf{\begin{tabular}[c]{@{}l@{}}Prefix-length buckets\\ Aggregate Encoding\end{tabular}}}} &
  LIME &
  0.82 &
  0.50 &
  0.57 \\ \cline{2-5} 
\multicolumn{1}{|l|}{} & SHAP & {\ul \textbf{1.00}} & {\ul \textbf{1.00}} & {\ul \textbf{1.00}} \\ \hline
\multicolumn{1}{|l|}{\multirow{2}{*}{\textbf{\begin{tabular}[c]{@{}l@{}}Prefix-length buckets\\ Index-based encoding\end{tabular}}}} &
  LIME &
  0.61 &
  0.33 &
  0.24 \\ \cline{2-5} 
\multicolumn{1}{|l|}{} & SHAP & {\ul \textbf{1.00}} & {\ul \textbf{1.00}} & {\ul \textbf{1.00}} \\ \hline
\end{tabular}
}
\end{table}

\begin{table}[b!]
\centering
\caption{Overall stability by weight results for each dataset}
\vspace*{-.25\baselineskip}
\label{tab:weight-stability-results}
\scriptsize{
\begin{tabular}{ll|c|c|c|}
\cline{3-5}
 &
   &
  \multicolumn{1}{l|}{\textbf{Production}} &
  \multicolumn{1}{l|}{\textbf{Sepsis Cases}} &
  \multicolumn{1}{l|}{\textbf{BPIC 2012}} \\ \hline
\multicolumn{1}{|l|}{\multirow{2}{*}{\textbf{\begin{tabular}[c]{@{}l@{}}Single bucket\\ Aggregate Encoding\end{tabular}}}} &
  LIME &
  0.95 &
  0.36 &
  0.85 \\ \cline{2-5} 
\multicolumn{1}{|l|}{} & SHAP & {\ul \textbf{1.00}} & {\ul \textbf{1.00}} & {\ul \textbf{1.00}} \\ \hline
\multicolumn{1}{|l|}{\multirow{2}{*}{\textbf{\begin{tabular}[c]{@{}l@{}}Prefix-length buckets\\ Aggregate Encoding\end{tabular}}}} &
  LIME &
  0.89 &
  0.63 &
  0.67 \\ \cline{2-5} 
\multicolumn{1}{|l|}{} & SHAP & {\ul \textbf{1.00}} & {\ul \textbf{1.00}} & {\ul \textbf{1.00}} \\ \hline
\multicolumn{1}{|l|}{\multirow{2}{*}{\textbf{\begin{tabular}[c]{@{}l@{}}Prefix-length buckets\\ Index-based encoding\end{tabular}}}} &
  LIME &
  0.80 &
  -0.04 &
  0.30 \\ \cline{2-5} 
\multicolumn{1}{|l|}{} & SHAP & {\ul \textbf{1.00}} & {\ul \textbf{1.00}} & {\ul \textbf{1.00}} \\ \hline
\end{tabular}
}
\end{table}

\subsubsection{Stability}  
Overall, SHAP is significantly more stable than LIME, with perfect stability by subset (see Table~\ref{tab:subset-stability-results}) and stability by weight (see Table~\ref{tab:weight-stability-results}). 
LIME is least stable when using index-based encoding and most stable when using aggregate encoding. For BPIC2012, single bucketing produces the best results, while prefix-length bucketing yields the best result for Sepsis Cases. LIME is almost always most stable for the Production dataset.  

\subsubsection{Fidelity}
Overall, fidelity was low-to-moderate for both explainable methods as shown in Table~\ref{tab:fidelity-results}. SHAP is once again the better performer, though generally by small margins. Both interpreters were generally less faithful when single buckets and aggregate encoding are used. Prefix-length bucketing and aggregate encoding produced the best results for the BPIC2012 event log, while prefix-length bucketing and index-based encoding produced the best results for the Sepsis dataset. LIME performed better with the former on the Production dataset, but SHAP performance was on par for both. 

\begin{table}[h!]
\vspace*{-.75\baselineskip}
\centering
\caption{Overall fidelity results for each dataset}
\vspace*{-.25\baselineskip}
\label{tab:fidelity-results}
\scriptsize{
\begin{tabular}{ll|r|r|r|}
\cline{3-5}
 &
   &
  \multicolumn{1}{l|}{\textbf{Production}} &
  \multicolumn{1}{l|}{\textbf{Sepsis Cases}} &
  \multicolumn{1}{l|}{\textbf{BPIC 2012}} \\ \hline
\multicolumn{1}{|l|}{\multirow{2}{*}{\textbf{\begin{tabular}[c]{@{}l@{}}Single bucket\\ Aggregate Encoding\end{tabular}}}} &
  LIME &
  0.26 &
  0.36 &
  0.37 \\ \cline{2-5} 
\multicolumn{1}{|l|}{} & SHAP & {\ul \textbf{0.27}} & {\ul \textbf{0.46}} & {\ul \textbf{0.41}} \\ \hline
\multicolumn{1}{|l|}{\multirow{2}{*}{\textbf{\begin{tabular}[c]{@{}l@{}}Prefix-length buckets\\ Aggregate Encoding\end{tabular}}}} &
  LIME &
  0.47 &
  0.37 &
  0.38 \\ \cline{2-5} 
\multicolumn{1}{|l|}{} & SHAP & {\ul \textbf{0.51}} & {\ul \textbf{0.49}} & {\ul \textbf{0.42}} \\ \hline
\multicolumn{1}{|l|}{\multirow{2}{*}{\textbf{\begin{tabular}[c]{@{}l@{}}Prefix-length buckets\\ Index-based encoding\end{tabular}}}} &
  LIME &
  0.36 &
  0.51 &
  0.32 \\ \cline{2-5} 
\multicolumn{1}{|l|}{} & SHAP & {\ul \textbf{0.51}} & {\ul \textbf{0.56}} & {\ul \textbf{0.4}}  \\ \hline
\end{tabular}
}
\vspace*{-2\baselineskip}
\end{table}

\subsection{Analysis and Findings}
\label{analysis}

\begin{figure}[h!]
\centering
\makebox[0pt] {
\includegraphics[scale=0.425]{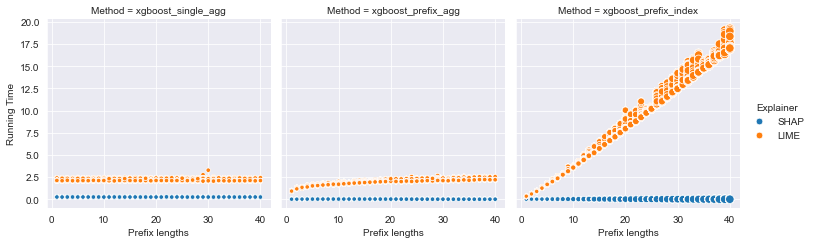}
}
\vspace*{-.5\baselineskip}
\caption{Time taken to create explanations for the BPIC2012 dataset (marker size indicates size of feature vector)}
\label{fig:timing}
\end{figure}

The running time of a predictive model based on each method combination was recorded for each interpreter (or explainer). 
SHAP was the fastest, taking less than a second per explanation. Time taken by LIME was reasonably consistent when using aggregate encoding, but when using index-based encoding the running time increased proportionally to prefix length (see Fig.~\ref{fig:timing}). A closer analysis suggests that the performance of LIME and SHAP is dependent on the characteristics of the datasets as well as the bucketing and encoding methods. The running time of LIME, in particular, is linked to feature vector length, increasing with the length of the feature vector input. This is especially apparent when using index-based encoding, where the feature vector size increases for each additional prefix, but running time remains consistent for aggregate encoding, where feature vector lengths are relatively consistent. SHAP is better able to handle this complexity, generally producing explanations in less than one second. 

To understand how instance-level characteristics and results affect the quality of explanations, we unfold the overall (dataset-level) results of evaluation measures into instance-level results (calculated according to the three equations defined in Sect.~\ref{metrics}). In particular, we can examine the effects of prefix length and the initial prediction probability provided by the black box on the result, along with the bucketing and encoding methods used. 

\begin{figure}[b!]
\vspace*{-.5\baselineskip}
\centering
\subfloat[Stability by Subset]{
\includegraphics[scale=0.35]{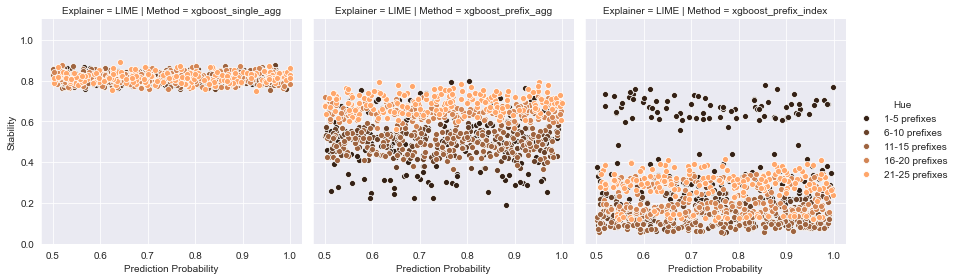}
}
\hspace{1pt}
\subfloat[Stability by Weight]{
\includegraphics[scale=0.35]{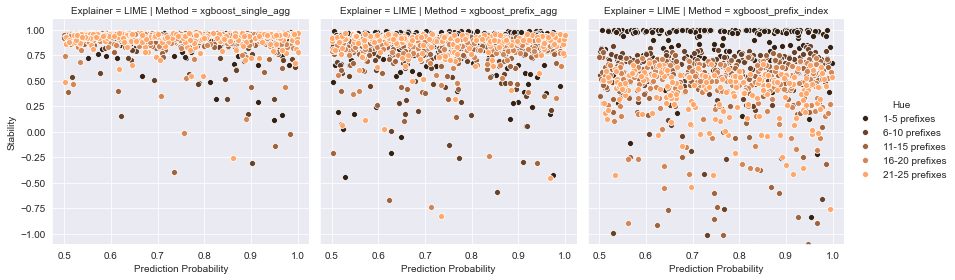}
}
\caption{Results of stability by subset (a) and stability by weight (excl. outliers) (b) of LIME over prediction probability and prefix length for the BPIC2012 dataset}
\vspace*{-.75\baselineskip}
\label{fig:bpic2012-stability}
\end{figure}

\vspace*{-.5\baselineskip}
\subsubsection{Stability}
As with running time, instance-level characteristics affect the stability of LIME. SHAP is almost perfectly stable for all classifiers, and there is no further information to be found when exploring the stability results of SHAP. But closer analysis suggests that the stability of LIME is, once again, dependent on the size of the feature vector (see Fig.~\ref{fig:bpic2012-stability}). This is likely because, to generate a surrogate model, LIME randomly samples inputs for the surrogate model from the neighbourhood of the input $x$. As the size of the feature vector increases, LIME's sampling efficiency and the representativeness of the sample decreases, resulting in surrogate models changing for each explanation, causing instability~\cite{Visani2020}. This problem is less visible when measuring stability by weight, but does exist and is most apparent when using index-based encoding. Oddly, the Production dataset is more stable than Sepsis when index-based encoding is used, but has longer prefix lengths. This is likely because of the dataset's sparsity, especially since a single black box is trained for each prefix-length. Once LIME has generated a set of input points $X'$ based on $x$, it uses the black box to create a set of labels $Y(x')$ for each of these inputs and uses this as ground truth to fit a surrogate model. It appears the underlying classifiers for the Production dataset when using index-based encoding and prefix-length bucketing are overfit (testing set accuracy is 61.8\% and training set accuracy is 92.5\%), resulting in a poorly-fit (likely under-fit) surrogate model as the generated labels for the new instances are likely near-random.

\begin{figure}[htbp]
\centering
\vspace*{-0.75\baselineskip}
\subfloat[Production]{
\includegraphics[scale=0.35]{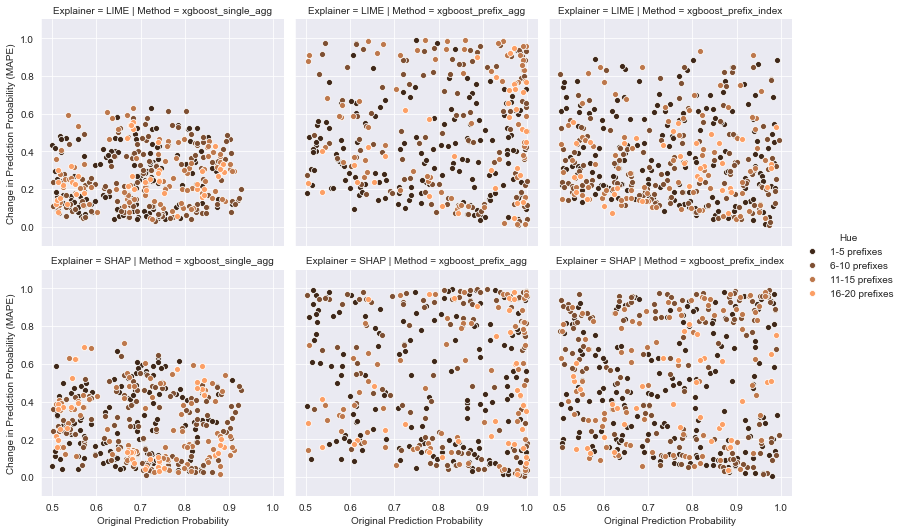}
}
\vspace*{-.5\baselineskip}
\hspace{1pt}
\subfloat[Sepsis Cases]{
\includegraphics[scale=0.35]{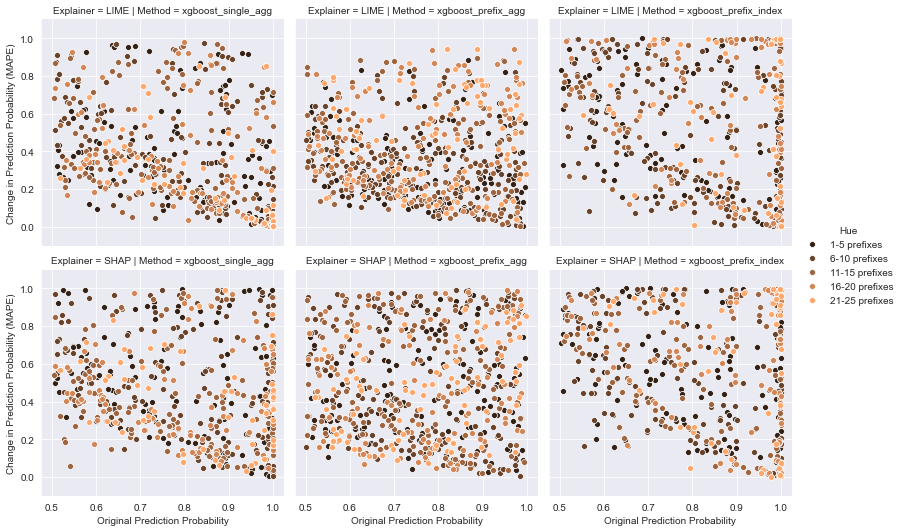}
}
\vspace*{-.5\baselineskip}
\hspace{1pt}
\subfloat[BPIC2012]{
\includegraphics[scale=0.35]{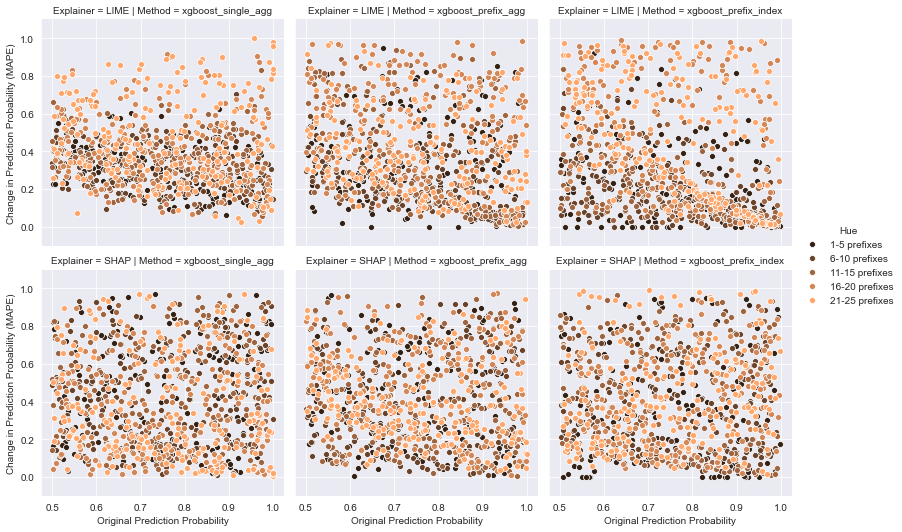}
}
\vspace*{-.25\baselineskip}
\caption{Fidelity results over original prediction probability and prefix length of LIME and SHAP, respectively, for Production (a), Sepsis Cases (b) and BPIC2012 (c)}
\vspace*{-1.5\baselineskip}
\label{fig:fid-all}
\end{figure}

\vspace*{-.25\baselineskip}
\subsubsection{Fidelity}
LIME and SHAP are almost comparable when evaluating fidelity. In many cases, such as in the Production dataset (excluding when single bucketing and aggregate encoding are used) or with SHAP for the BPIC2012 dataset (see Fig.~\ref{fig:fid-all}), the faithfulness of explanations varies across instances in an almost uniform distribution. This suggests that only some explanations are faithful, but there appears to be no pattern or trend of faithfulness with regards to prefix length, encoding method, bucketing method or the initial prediction probability. As noted, both interpreters are least faithful when single bucketing and aggregate encoding are used together, but this combination has been identified as the most effective in ensuring black box accuracy~\cite{Teinemaa2019}. As such, it may be difficult to apply these interpretation methods in practice, unless other bucketing and encoding techniques are reasonably accurate for that dataset and black box. However, it is important to note that, while the interpreters are more faithful for other combinations, results are generally quite low across all combinations. A closer examination of higher-performing combinations, such as the Sepsis Cases dataset with index-based encoding or the Production dataset with prefix-length bucketing and aggregate encoding, shows that, for the most part, changes in prediction probability are quite small. Here, it appears that some instances with more faithful explanations have raised the overall fidelity score. 

\begin{figure}[h!]
\centering
\makebox[0pt]{
\subfloat[\tiny Single Bucket,\\ Aggregate Encoding]{
\includegraphics[scale=0.3]{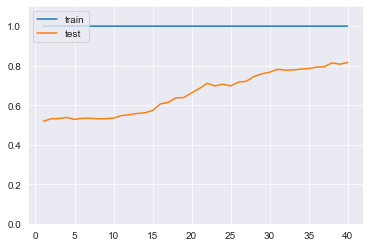}
}
\subfloat[\tiny Prefix-length Bucket,\\ Aggregate Encoding]{
\includegraphics[scale=0.3]{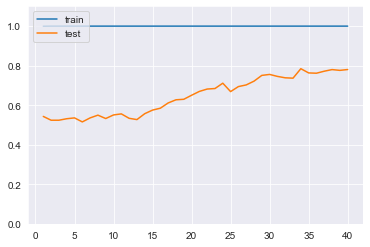}
}
\subfloat[\tiny Prefix-length Bucket,\\ Index-based Encoding]{
\includegraphics[scale=0.3]{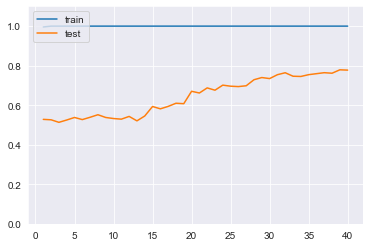}
}
}
\caption{Accuracy of predictive models at each prefix length for the BPIC2012 dataset}
\label{fig:bpic2012-acc}
\end{figure}

There is generally no link between prefix length and fidelity, except with BPIC2012 (most noticeable in LIME), where a higher prefix length generally results in a more faithful explanation. The large size of the BPIC2012 dataset (at least in comparison to the other two) has resulted in more reasonable black box accuracy at the higher prefix lengths (see Fig.~\ref{fig:bpic2012-acc}), which in turn appears to have ensured that LIME's surrogate models better fit the data. Interestingly, there is no pattern between stability and fidelity. It may be expected that when stability is poor, fidelity will also be poor given the variation in explanations. Generating multiple explanations to choose the most common features appears to have mitigated this. This is particularly clear in the Sepsis Cases dataset, which had the poorest stability results, but performs comparably with the other datasets for fidelity. It is possible that this approach can be used to mitigate stability issues in practice (provided that the interpretation method is efficient enough to make this feasible for timely decision-making).

The Sepsis Cases dataset presents an interesting pattern when index-based encoding used. From the results, it becomes apparent that explanations from both LIME and SHAP are more uniformly faithful for instances closer to the decision boundary (i.e. when the initial prediction probability is closer to 0.5), and less precise in fidelity when moving away from the decision boundary. This is likely, once again, a combination of data sparsity in each bucket (caused by the down-sampling) and relatively long feature vector lengths (in comparison to aggregate encoding) resulting in an under-fit black box model (training accuracy and testing accuracy are both low at 67.1\% and 45.8\% respectively) that make near-random predictions when unfamiliar instances are presented. 
\section{Conclusion}\label{concl}
As black box models are often applied in predictive process analytics, explainability becomes necessary to help understand why certain predictions are made by the underlying predictive models.  
To determine the quality of explanations, we have proposed functionally-grounded evaluation metrics and used them to assess explainable methods in process predictive analytics. 
The evaluation metrics presented in this paper are not limited to the prediction problems, predictive models and explainable methods explored. 
They can be extended to include further measures, such as measures specific to other aspects of quality, user-oriented measures, in future. 
As a next step, one interesting area to explore the support of explainability is prescriptive process monitoring, in which process recommendations are generated in addition to process predictions~\cite{Mehdiyev2020}. Another possible avenue is to evaluate explainability for other prominent predictive methods applied in predictive process analytics, such as LSTM~\cite{Verenich2019}. 
Future work exploring such extensions may be beneficial in identifying appropriate and/or suitable explainable methods for preditive process analytics. 

%
%
%
\bibliographystyle{splncs04}
\bibliography{6.references}

\end{document}